\documentclass[letterpaper, 10 pt, conference]{ieeeconf}

\IEEEoverridecommandlockouts                              

\usepackage[noformatting,nogeometry]{stdcommands}
\usepackage{graphicx}
\usepackage{dsfont} 
\usepackage{amsmath}
\usepackage{amssymb}
\usepackage{url}
\usepackage[utf8]{inputenc} 
\usepackage[format=hang,singlelinecheck=false,caption=false,font=footnotesize]{subfig}
\usepackage{changes}
\usepackage[ruled,linesnumbered]{algorithm2e}
\usepackage{scalerel}
\usepackage[noend]{algpseudocode}
\usepackage[english]{babel}
\usepackage{mathrsfs}
\usepackage{subfig}
\usepackage{caption}
\usepackage{comment}
\usepackage{multirow}

\title{\LARGE \bf Data Driven Prediction Architecture for Autonomous Driving and its Application on Apollo Platform}

\begin{document}

\author{%
    \parbox{\linewidth}{\centering
      Kecheng Xu, Xiangquan Xiao, Jinghao Miao, Qi Luo$^{1}$
  }%
  \thanks{$^1$ Corresponding author}
  \thanks{All Authors are with Baidu USA LLC,
        1195 Bordeaux Drive, Sunnyvale, CA 94089
        {\tt\small xukecheng@baidu.com, xiaoxiangquan@baidu.com,  miaojinghao@baidu.com,luoqi06@baidu.com}}%
  \thanks{This paper has been accepted by the 31st IEEE Intelligent Vehicles Symposium (2020)}
}

\markboth{}{}

\maketitle
\thispagestyle{empty}
\pagestyle{empty}

\begin{abstract}

Autonomous Driving vehicles (ADV) are on road with large scales. For safe and efficient operations, ADVs must be able to predict the future states and iterative with road entities in complex, real-world driving scenarios. How to migrate a well-trained prediction model from one geo-fenced area to another is essential in scaling the ADV operation and is difficult most of the time since the terrains, traffic rules, entities distributions,  driving/walking patterns would be largely different in different geo-fenced operation areas. In this paper, we introduce a highly automated learning-based prediction model pipeline, which has been deployed on Baidu Apollo self-driving platform, to support different prediction learning sub-modules' data annotation, feature extraction, model training/tuning and deployment. This pipeline is completely automatic without any human intervention and shows an up to 400\% efficiency increase in parameter tuning, when deployed at scale in different scenarios across nations.
\end{abstract}

\IEEEpeerreviewmaketitle


\section{Introduction}
%
%
%
%

\subsection{Motivation}
\label{subsec:motivation}
In order to fully autonomous driving in complex scenarios, comprehensive understanding and accurate prediction of driving entities' future states are crucial. Researchers from both academia and industry have been extensively studied the prediction techniques. However, rarely researchers discussed the challenges in scaling of the prediction models to different scenarios and entity types in real world application. On the other hand, Apollo platform~\cite{apollo}, has the challenge of running in different scenarios across the nations and on different vehicle platforms~\cite{apollo_2019}. So a main motivation of this work is a combined onboard-offboard and combined end-to-end prediction pipeline that helps scale the autonomous vehicle testing and operating to different scenarios, different traffic rule, and different vehicle platforms with enhanced efficiency and reduced cost.

\subsection{Related Work}
\label{subsection:related_work}
The prediction problems aim to estimate the entities future states (position, heading, velocity, acceleration etc.) in the next few seconds, and can usually be solved via two different approaches: One is~\emph{Direct trajectory prediction}: the model directly output the entities' future trajectories in discrete point format. Another is~\emph{Entities' intention prediction + post trajectory generation}: the model would only output entities' intentions with probabilities (change lane/no change, intersection exist lane, roundabout exit/no exit.) with an sampling/optimization based trajectory generator.

\subsubsection{prediction of intention + post trajectory generation}
\label{subsection:indirection_prediction}
Early works in ADV prediction usually use Kalman filter (KF) and its alternatives to estimate and propagate the entity future states~\cite{cosgun2017towards}, and Gaussian Process (GP) for human dynamic modeling~\cite{wang2007gaussian}.  While these approaches usually work well in simple short-term horizon, they generally fail to encoder environment context (such as road topology, traffic rules) thus downgrade in performance in complex environment.

Alternatively, some works formulate this problem with Partially-Observable Markov Decision Process (POMDP)~\cite{kitani2012activity} or Hidden Markov Model (HMM)~\cite{deo2018would} followed by inverse optimal control. Recently, Researcher also try combine the RNN based high-level policy anticipate with low level non-linear optimization based trajectory generation~\cite{8793568}. We call these \emph{indirect prediction trajectory generation} in following chapters.

\subsubsection{prediction algorithm with direct trajectory generation}
\label{subsection:prediction_algorithm}
The prediction models above usually model the ego-environment behavior in the environment with the assumption that obstacles behave independently of each other. Inspired by the successful applications of deep learning in computer vision and natural language processing, different deep learning approaches have been proposed to model both the ego-environment and obstacle-obstacle interactions within the environment. Researchers either implicitly model environment and these interactions via encoding, like~\cite{bansal2018chauffeurnet} and~\cite{chou2019predicting} do, or explicitly model the social interactions between obstacles by adding both the spatial and temporal information in LSTM based deep learning model structures~\cite{alahi2016social}\cite{vemula2018social}\cite{mohamed2020social}. Multi-modal prediction trajectories and trajectories' probabilities can also be added with softmax normalization~\cite{chai2019multipath}, using Variant Auto Encoding (VAE)~\cite{hong2019rules}, or Generative Adversarial Approach~\cite{li2019coordination}.  We call these \emph{direct prediction trajectory generation} in following chapters.

\subsection{Contributions}
Our main contributions are as follows:
\begin{enumerate}
	\item \textbf{Data Driven prediction architects aiming to support large scale operation}: This includes two major parts.
	
	\begin{enumerate}
	    \item \emph{Onboard part}: this is the part which is actually running inside the ADV and includes three major components: Message Pre-Processing, Model Inference and Trajectory Post-Processing.
	    \item \emph{Offboard part}: this is the part which does not run on the ADV, but instead running on the data-pipeline, this includes 5 components: Automatic Data Annotation, Feature Extraction, Model Training, Hyper-Parameter Auto-tuning and Result Evaluation. 
	\end{enumerate}
    
	\item \textbf{Indirect/direct prediction trajectories generation support}: We use two models to show that our pipeline architectures are flexible and completely automatic to support and accelerate scenario adaptions for both indirect and direct prediction generation: In Section~\ref{sec:semantic_lstm}, we use ~\emph{semantic map + LSTM} to show a complete pipeline for direct trajectory prediction through automatic data annotation and model structure improvement without any human intervention, yet achieve similar performance compared with model trained on man. In Section~\ref{sec:intention_prediction_and_post_trajectory_generation} we use~\emph{Intersection exit prediction model + Siamese auto tuning + post trajectory generation} as an example for indirect trajectory prediction pipeline. Siamese auto tuning increases efficiency by 400\%, avoiding manual parameter tuning process.
	
	\item \textbf{Real world application and open capability}: Following extensive onboard and offboard testings, this system has been deployed to several fleet of self-driving vehicles of different types in both China and US. Apollo platform may open the prediction data pipeline and model training service as we have done for other services.
	
\end{enumerate}

This paper is organized as: Section~\ref{sec:prediction_architecture} gives the introduction of prediction onboard and offboard components, Section~\ref{sec:semantic_lstm} and~\ref{sec:intention_prediction_and_post_trajectory_generation} give two examples of direct and indirect prediction trajectory generation process, and how they can be improved via our pipeline efficiency, Section~\ref{sec:conclusion} gives the conclusion remarks and future work.
\section{Prediction Module Architecture in Apollo Platform}
\label{sec:prediction_architecture}
This section gives an introduction of prediction modules architecture in Apollo platform including both the onboard (on vehicle) components as well as the offboard (in data pipeline) components. 

\subsection{Onboard Architecture}
\label{sec:onboard_architecture}
\begin{figure*}[!htb]
  \centering
  \includegraphics[width=16cm, height=10cm]{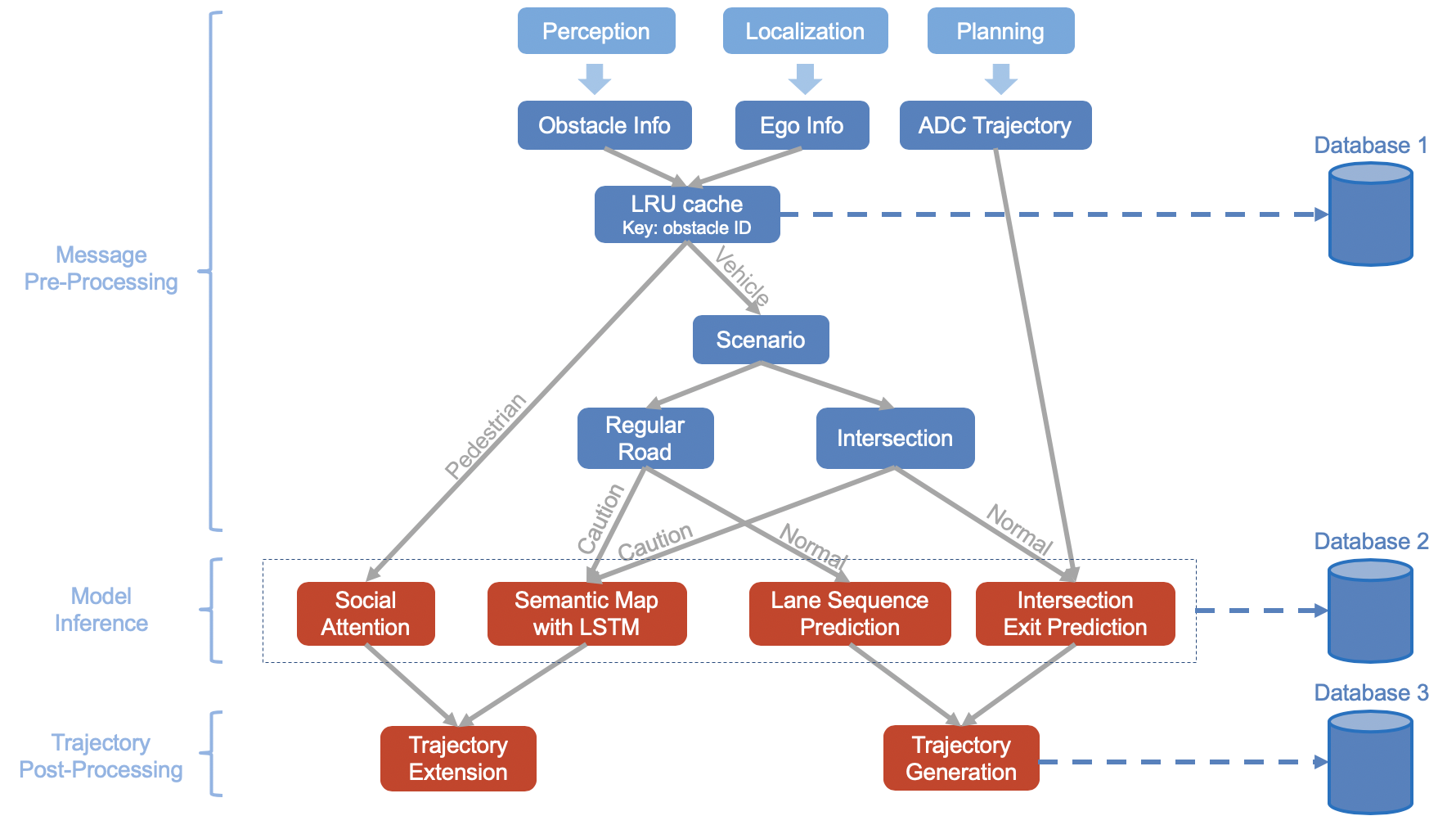}
  \caption{Onboard architecture of the prediction module on Apollo autonomous driving open-source platform.}
  \label{fig:online_architecture}
\end{figure*}

In this section, we introduce the onboard architecture of the prediction module on Apollo autonomous driving open-source platform. This architecture supports multiple obstacle categories, multiple scenarios and multiple levels of obstacle attention priorities. The structure of onboard workflow is shown in Figure \ref{fig:online_architecture}.

\subsubsection{Message Pre-Processing}
\label{subsec:message_preprocessing}




As shown in Figure~\ref{fig:online_architecture}, \emph{message pre-processing} has two objectives:

    \begin{itemize}
	        \item Merge localization/perception output to prepare environment context for machine learning models, and dump these information for future training/evaluation purpose.
	        \item Scenario selection (intersection and regular road), obstacle prioritization (caution and normal) based on environment context and previous planning trajectories. 
    \end{itemize}
    
\subsubsection{Model Inference}
\label{subsec:model_inference}

Once the scenario, obstacle type and priority have been determined in~\emph{Message Pre-Processing}, this sub-module selects the corresponding model for inference.  

We take pedestrians and vehicles as examples. For a pedestrian, we apply social-attention model to predict pedestrian's future trajectory for next four seconds. For a vehicle, if it is of prioritized in caution level, we apply a \emph{semantic map + LSTM} model with direct prediction trajectory output with uncertainty information. If it is of prioritized in normal level, we apply~\emph{lane sequence model} or~\emph{intersection exit model} to predict vehicle's intention like which lane or exit the obstacle vehicle will chose next, as shown in Figure~\ref{fig:online_architecture}.


\subsubsection{Trajectory Generation or Trajectory Extension}
\label{subsec:trajectory_generation_and_extension}
This sub-module serves for as model post-processing with two main objectives: 

\begin{itemize}
\item For pedestrians or caution-prioritized vehicles, we~\emph{extend} the direct generated trajectories with KF with different kinodynamic modeling up to 8s.  
\item For normal-prioritized vehicle, we take the prediction intention and~\emph{generate} the 8s prediction trajectories via a sampling-based method with cost autotuned for different scenarios. More details would be discussed in Section~\ref{sec:intention_prediction_and_post_trajectory_generation}.
\end{itemize}

\subsection{Offboard Architecture}
\label{sec:offboard_architecture}

\begin{figure*}[!htb]
  \centering
  \includegraphics[width=16cm, height=10cm]{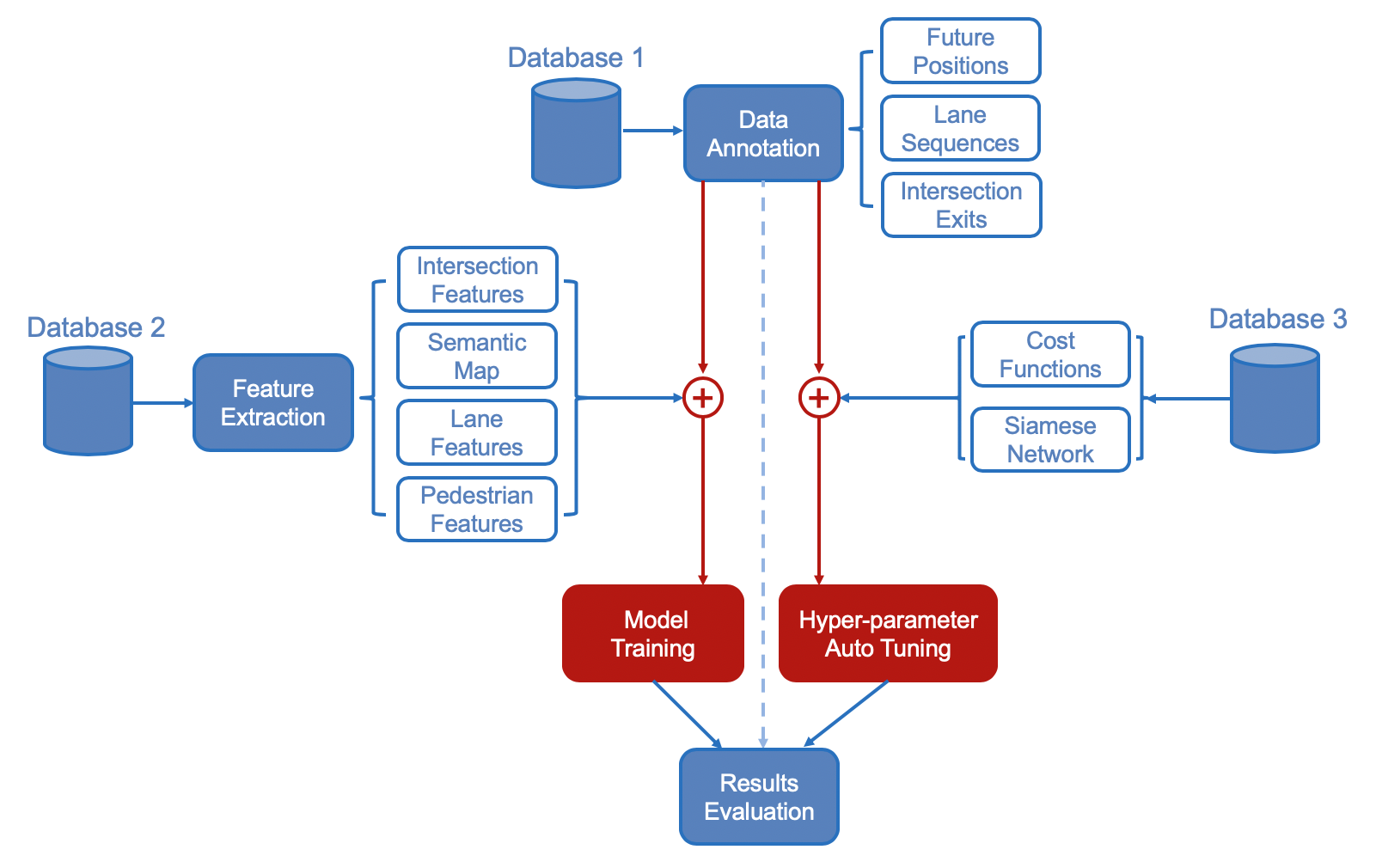}
  \caption{Offboard architecture of the prediction module on Apollo autonomous driving open-source platform.}
  \label{fig:offline_architecture}
\end{figure*}

In this section, we introduce the offboard architecture support data processing and model training. As shown in Figure \ref{fig:offline_architecture}, this architecture includes 1) automatic data annotation, 2) feature extraction and model training 3) auto tuning and 4) results evaluation.

\subsubsection{Automatic Data Annotation}
\label{subsec:data_labeling}
This sub-module takes time stamped perception and localization information, map topology information as well as traffic rules dumped from onboard Database I to generate "ground truth" labels. Depending on specific task/ML model needs, the annotation labels can includes but not limited to: future positions, lane sequence labeling, and intersection exit labeling, etc.

\subsubsection{Feature Extraction and Model Training}
\label{subsec:feature_extraction}
This sub-module takes the features for different models from Database 2 with the key of the combination of road test ID, obstacle ID and timestamp. We also have a sub-key to distinguish the features for different models including intersection features, semantic map, lane sequence features and pedestrian historical movement features. These key-value system make feature query, combination more easier for different model training tasks.


\subsubsection{Auto Tuning}
\label{subsec:auto_tuning}

Auto tune sub-module here only applicable to the ~\emph{indirect prediction trajectory generation}, where after we get the intention prediction results, we need to use either graphical search, curve fitting or optimization based method to generate prediction trajectory afterwards. In all three methods, cost tuning are needed for different scenarios. This cost tuning can be done via simple logistic regression to Inverse Reinforcement Learning (IRL) framework~\cite{kober2013reinforcement} or Bayesian based methods~\cite{neumann2019data}, in section~\ref{sec:intention_prediction_and_post_trajectory_generation}, we use Siamese network as an example search for the optimal cost in post trajectory generation.

\subsubsection{Results Evaluation}
\label{subsec:result_evaluation}
Results evaluation task measures how accurate obstacle's predicted trajectories are, compared with its actual future trajectories. Common trajectory evaluation metrics include Average Displacement Error (ADE), Final Displacement Error (FDE), other evaluation metrics may include  Dynamic Time Warping (DTM)~\cite{senin2008dynamic}, CLEAR-MOTA~\cite{bernardin2008evaluating} or Weighted Brier Score~\cite{zhan2018towards}. In following sections, we will use ADE and FDE for results comparison with peers. 


\section{Prediction Model based on Semantic Map and LSTM}
\label{sec:semantic_lstm}

In this section, we use our semantic map encoding + LSTM as an example to show the efficiency increase for a typical direct prediction trajectory generation workflow, including both offboard pipeline and onboard components that:

\begin{enumerate}
	    \item \emph{Efficiency improvement}: We give an example usage of the completely automatic pipeline for semantic map encoding avoiding any human intervention, thus support easy scenario extension and scaled ADV fleet deployment.
	    \item \emph{Performance comparable}: We show that through prediction model structure redesign we achieve similar performance as in table~\ref{table:model_comparison}.
	\end{enumerate}

\subsection{Automatic Annotation and Semantic Map Encoding}
\label{subsec:semantic_map}

\begin{figure}[!htb]
  \centering
  \includegraphics[width=4cm, height=4cm]{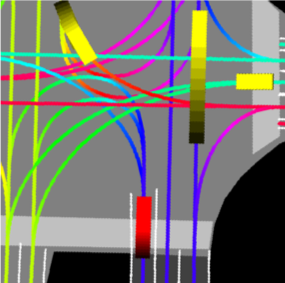}
  \caption{Semantic map for the target vehicle.}
  \label{fig:semantic_map_target}
\end{figure}

We use a semantic map similar to~\cite{DBLP:journals/corr/abs-1808-05819} to encode the dynamic context in environment and its past history, as shown in Figure~\ref{fig:semantic_map_target}. The semantic map encode a 40$\times$40 square meters environment (includes all road entities, map topology as well as traffic rules) into a 400$\times$400 pixels image centered with target vehicle. The target vehicle is marked red, while other entities (vehicles, bicycles, pedestrians) are marked yellow, with their historical trajectories marked with darker color. Note that the "target vehicle" here refers not necessarily only our own ADV, but also can be the obstacle vehicles on road. We use timestamped perception results as automatic annotation for all road entities' histories and create a semantic map for each vehicle. 


\subsection{Model Structure and Loss Function}
\label{subsec:semantic_lstm_method}

\begin{figure*}[!htb]
  \centering
  \includegraphics[width=14cm, height=6cm]{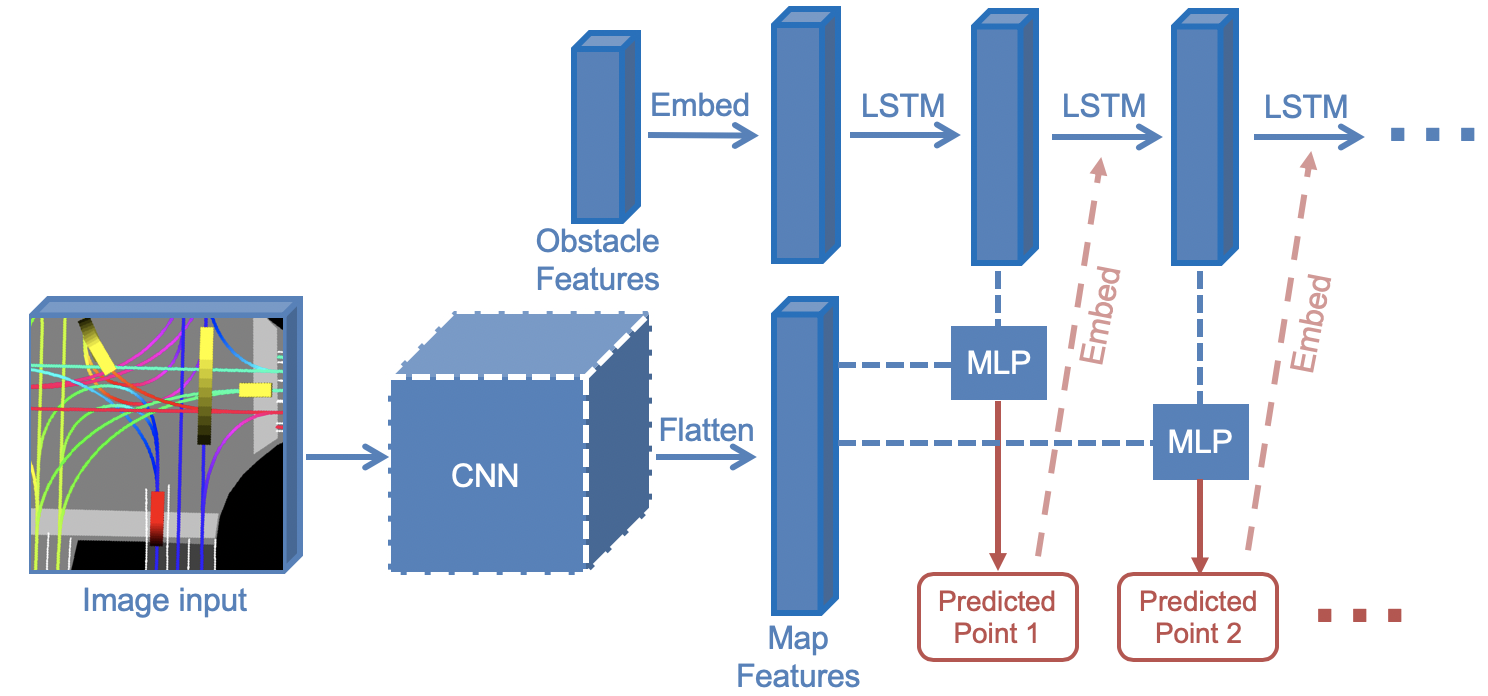}
  \caption{Structure of semantic map + LSTM model.}
  \label{fig:model_structure}
\end{figure*}

The model combines LSTM, CNN, and MLP. The input of CNN part is the semantic map described in Section \ref{subsec:semantic_map}. The output of of CNN is a feature vector. In the final version of the model, MobileNet-v2 is chosen in the CNN part. And the output is the flatten vector from the second from last layer.

The LSTM sequence consist of historical part and prediction part. In the historical part we embed the relative coordinates to its current position and then apply the embedded feature to update the LSTM hidden state. In the prediction part, we concatenate LSTM hidden state and the output feature vector of CNN, then pass the concatenation into an MLP to get a predicted position point relative to the current position. Then we use the predicted position to get the next embedding feature, updating the LSTM hidden state which will be used to predict the next future relative position. The procedure is shown in Figure \ref{fig:model_structure}. We continue this procedure until we get 30 future relative positions which stand for a 3-second predicted trajectory because the time resolution of predicted trajectory is 0.1 second.

We have two loss functions. One is mean squared error (MSE) described in Equation \eqref{eq:mse}.
\begin{equation}
  \text{MSE} = \frac{1}{N}\sum\limits_{i = 1}^{N}
        [(x_i^{pred} - x_i^{true})^2 + (y_i^{pred} - y_i^{true})^2]
  \label{eq:mse}
\end{equation}
where, $N$ is the number of predicted trajectory points. With the MSE loss, the model output trajectory points for three seconds. The other loss function is negative log likelihood (NLL) based on bi-variate Gaussian distribution as Equation \eqref{eq:gaussian_loss}
\begin{equation}
  \text{Loss} = -\frac{1}{N}\sum\limits_{i = 1}^{N}\log{P}
  \label{eq:gaussian_loss}
\end{equation}
where $P$ is a bi-variate probability density function with mean $(\mu_x, \mu_y)$, and covariate matrix $\begin{pmatrix}
\sigma_x^2 & \rho\sigma_x\sigma_y\\
\rho\sigma_x\sigma_y & \sigma_y^2
\end{pmatrix}$. With the NLL loss, the model outputs the Gaussian distributions of the future trajectory points for three seconds.

\begin{figure}[!htb]
  \centering
  \includegraphics[width=0.45\textwidth]{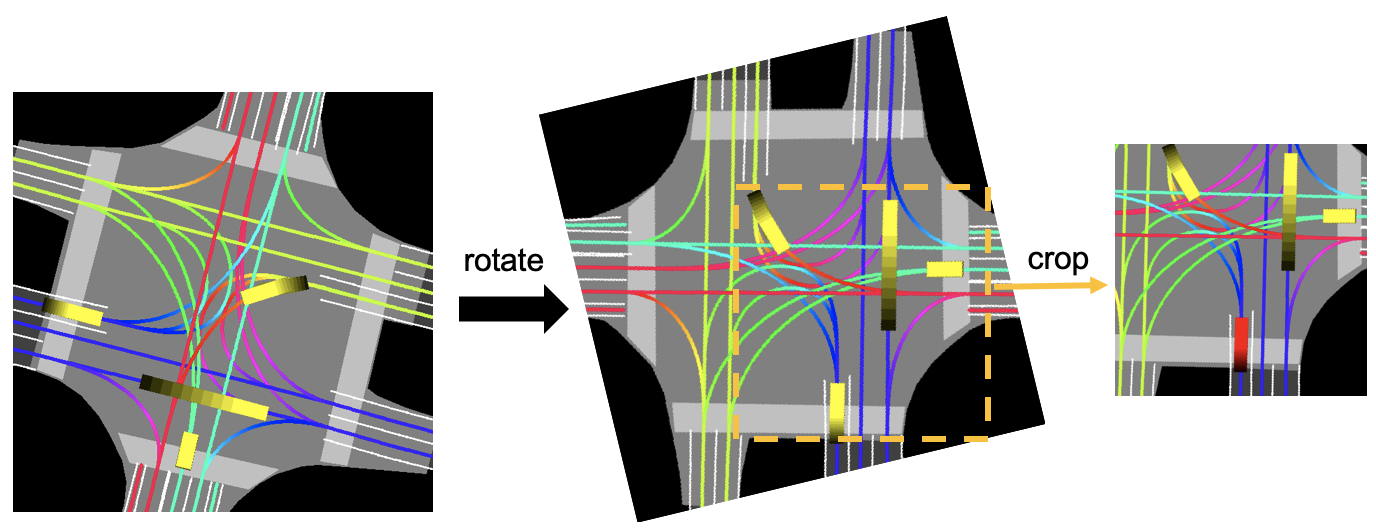}
  \caption{Process to crop small image for each obstacle.}
  \label{fig:crop}
\end{figure}

The model is trained from 1000 hours' urban driving traffic data by Lincoln MKZs equipped with Velodyne HDL-64E LiDar. The software for obstacle detection and localization was Apollo 5.0 (https://github.com/ApolloAuto/apollo/tree/r5.0.0). For each surrounding vehicle, we crop a small image which stands for its local area as Figure \ref{fig:crop}. In each small image, the corresponding obstacle's heading is upside, and its current and historical polygons are marked as red. This small image is the input of the CNN part of the model shown in Figure \ref{fig:model_structure}.

\subsection{Evaluation and Result Comparison}
\label{subsec:semantic_evaluation}
We compared our model performance with the state-of-art from industry and show the results in Table \ref{table:model_comparison}. We compared the results of our models with peers' on auto-annotated Apollo dataset and showed that our model out performed in ADE and FDE for both 1s and 3s.  And achieved similar state-of-art 3s ADE results (0.77m vs 0.71m) with peer's best performance on their internal dataset. Due to commercial restriction, We can only  selectively present results in Sunnyvale, CA and San Mateo, CA as a demonstration to show that our model and pipeline achieved similar results, regardless of different driving patterns.  But the system performance has been validated under different geo-fenced areas across countries (China and United States).

We also investigated the robustness of \emph{semantic map + LSTM + uncertainty} model in different environments. Table \ref{table:semantic_lstm_stability} shows a pretty robust performance for different test environments including going straight, turning left, turning right and changing lane.

\begin{table*}[!htb]
\centering
\begin{tabular}{lclccccccc}
\hline
\hline
  &    &   &         \multicolumn{4}{c}{Apollo Data}        &      \multicolumn{3}{c}{Others' Internal Data}     \\
Team & Scenario & Model    & ADE(1s) & FDE(1s) & ADE(3s) & FDE(3s) & ADE(1s) & ADE(3s) & ADE(5s) \\
\hline
\multirow{6}{*}{Apollo} & \multirow{3}{*}{Sunnyvale} & LSTM    & 0.26m  & 0.48m   & 1.33m  & 3.34m & - & - & -  \\
& & Semantic map + LSTM & 0.23m & \textbf{0.37m} & \textbf{0.77m} & \textbf{1.85m} & - & - & -  \\
& & Semantic map + LSTM + uncertainty    & \textbf{0.22m}  & 0.38m   & 0.79m  & 1.93m & - & - & -  \\
\cline{2-10}
& \multirow{3}{*}{San Mateo} & LSTM & 0.26m & 0.51m & 1.35m & 3.41m & - & - & -  \\
& & Semantic map + LSTM & 0.24m & \textbf{0.39m}  & \textbf{0.79m}  & \textbf{1.91m} & - & - & -  \\
& & Semantic map + LSTM + uncertainty & \textbf{0.21m} & 0.40m & 0.80m  & 1.98m & - & - & -  \\
\hline
Uber & - & Semantic map + MLP \cite{DBLP:journals/corr/abs-1808-05819}  & 0.29m  & 0.51m   & 0.97m  & 2.33m &  & \textbf{0.71m} &  \\
ZooX & - & Semantic map + GMM + CVAE \cite{DBLP:journals/corr/abs-1906-08945}   &  -  & - & - & - & 0.44m & - & 2.99m \\
\hline\hline
\end{tabular}
\caption{Model performance comparison.}
\label{table:model_comparison}
\end{table*}

\begin{table}[!htb]
\begin{tabular}{lllll}
\hline
\hline
Behavior    & ADE(1s) & FDE(1s) & ADE(3s) & FDE(3s) \\
\hline
Straight    & 0.229m  & 0.371m   & 0.776m  & 1.894m \\
Turn Left   & 0.248m  & 0.385m   & 0.744m  & 1.718m \\
Turn Right  & 0.299m  & 0.432m   & 0.867m  & 2.049m \\
Change Lane & 0.261m  & 0.412m   & 0.787m  & 1.813m \\
\hline\hline
\end{tabular}
\caption{Stability analysis of semantic map + LSTM model.}
\label{table:semantic_lstm_stability}
\end{table}

\section{Intention Prediction and Post Trajectory Generation}
\label{sec:intention_prediction_and_post_trajectory_generation}

\begin{figure*}[!htb]
  \centering
  \includegraphics[width=12cm, height=10cm]{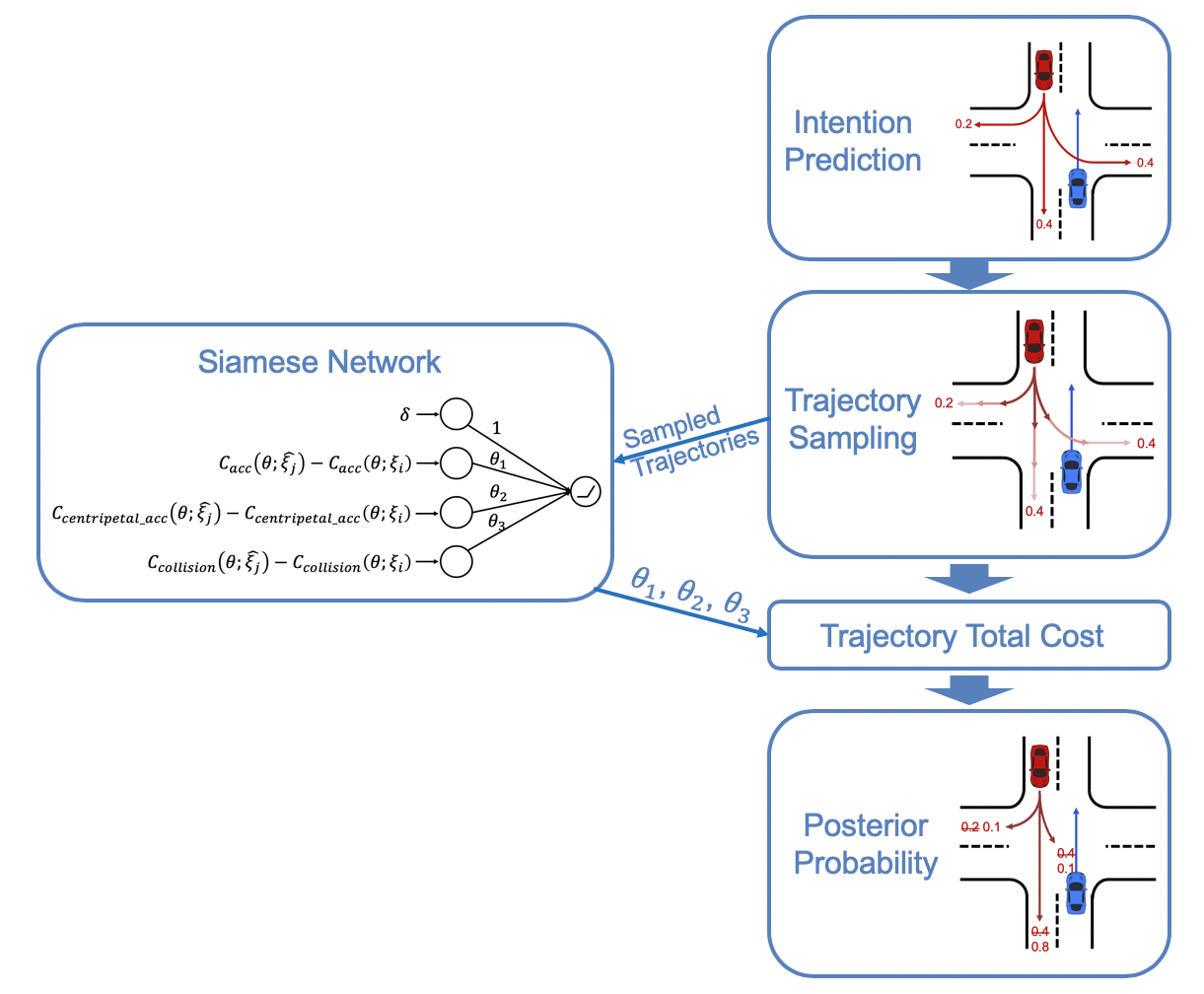}
  \caption{Workflow of intention prediction and post trajectory generation.}
  \label{fig:siamese}
\end{figure*}

We support different intention prediction model in Apollo platform and in this section, we will use Intersection MLP (shown in Figure~\ref{fig:siamese} as an example to show how the scenario adaption can be efficiently done in our prediction pipeline.

\label{subsec:siamese_network}

\subsection{MLP based Intention Prediction model for Intersection Exit}
\label{subsec:intention_prediction}
Figure \ref{fig:siamese} upper right part shows the intersection exit model for the red vehicle. The model takes obstacle's historical states (positions, headings, velocities) and all intersection exits' features (positions and headings) as inputs and outputs three intentions: going straight, turning left and turning right with probabilities of 0.4, 0.4 and 0.2 as priors in the post trajectory generation stage.

\subsection{Post Trajectory Generation}
\label{subsec:post_trajectory_generation}
Once we get the intention output, ~\emph{post trajectory generation } is the next submodule the complete prediction trajectories using a lattice-like sampling methods:

	\begin{enumerate}
	    \item Generate prediction paths for each intention via lane sequence search.
	    \item Along each path, sample different temporal profiles within vehicle physical limits.
	    \item Combine paths and temporal profiles to generate trajectories, and select the best trajectory as output according to trajectory posteriors distribution defined in Equation~\ref{eq:posterior}.  
	\end{enumerate}

Equation~\ref{eq:posterior} gives the trajectory posterior calculation. Where~\emph{prior} is output from MLP model, $Z$ is the normalization factor and $C$ is the trajectory total cost calculated in Equation~\ref{eq:cost_total}. 

\begin{equation}
    \textrm{Posterior} = \frac{1}{Z} \cdot \textrm{prior} \cdot e^{-C}
    \label{eq:posterior}
\end{equation}

The total cost in Equation~\ref{eq:cost_total} is weighted cost from different trajectory evaluation metrics, such as acceleration, centripetal acceleration and collision cost. $Z_1$ and $Z_2$ are the normalization terms, and $d_i$ is point-wise distance error between ego vehicle's previously planned trajectory and obstacles' predicted position. 
Note that $\theta_1$ to $\theta_3$ in this equation are the weights reflecting the driving patterns of road entities. And these are dramatically different in different geo-fenced area and operation time.

\begin{subequations}
  \label{eq:cost_total}
  \begin{align}
    &
    C = \theta_1 C_{acc} + \theta_2 C_{centripetal\_acc} + \theta_3 C_{collision} \\
    & \text{where: } \\
    & \hspace{2.0em} C_{acc} = \sum\limits_{i}a_i^2 \label{cost:acc}, \\
    & \hspace{2.0em} C_{centripetal\_acc} = \frac{1}{Z_1}\sum\limits_{i}(v_i^2\kappa_i)^2 \label{cost:centripetal_acc}, \\
    & \hspace{2.0em} C_{collision} = \frac{1}{Z_2}\sum\limits_{i}e^{-d_i^2}
  \end{align}
\end{subequations}



\subsection{Siamese Network For Efficiency Improvement}



Manual tuning weighs for Equation~\ref{eq:cost_total} is really low efficient and by no means makes large deployment possible, in order to support fleet deployment at scale in different geo-fenced areas, we introduce an auto-tune submodule to automatically find the optimal weights in different scenarios. We use Siamese network here as an example, but this can be of IRL or Bayesian based approach as well. The key thing is that those methods need to share the basic assumption that human trajectories (excluding the unsafe ones), are optimal in a statistical manner.  

The inputs of Siamese network are the sub-costs of the sampled trajectories and the ground-truth real trajectories. We use the weights for different costs in post trajectory generation after training the network.
We use $\xi$ to denote a sampled candidate trajectory, and $\hat\xi$ to denote a ground-truth real trajectory. We re-represent their cost function of as $C(\theta;\xi)$ and $C(\theta;\hat\xi)$. The objective function for the Siamese Network is designed as Equation~\eqref{eq:objective}.
\begin{equation}
    L = \sum\limits_{j=0}^{N}\sum\limits_{i=0}^M|C(\theta;\hat\xi_i)-C(\theta;\xi_j) + \delta|_+
    \label{eq:objective}
\end{equation}
where, $|\cdot|_+$ is the maximum between $\cdot$ and 0, $N$ is the number of data (an obstacle at a timestamp), $M$ is the number of sample candidate trajectories for each data. $\delta$ is marginal factor which is a small positive constant. We tried to solve the optimization problem to minimize the objective function defined in Equation~\eqref{eq:objective}. The marginal constant $\delta$ is necessary added to avoid getting all zero weights. Also, the relatively small value of $\delta$ can omit the affect of the sampled trajectories with costs much larger than the ground-truth trajectory cost, in which situation, $|C(\theta;\hat\xi_i)-C(\theta;\xi_j) + \delta|_+ = 0$.

Once we get the optimal weights, the onboard~\emph{post trajectory generation} submodule can use this to rank and choose optimal predicted trajectories for different geo-fenced areas with different driving patterns, traffic rules, etc. We show a roughly 400\% efficiency increase compared with manual parameter tuning when deployed in a new geo-fenced area. 



\section{Conclusion}
\label{sec:conclusion}
In this paper, we present a complete data driven prediction architecture including both the onboard part and offboard parts. We show this with two example how the direct/indirect prediction generation methods can benefit from the automatic data annotation, training process and hyper-parameter tuning to reduce the deployment effort across different scenarios.

\section*{Acknowledgment}
The authors would like to thank the Apollo Quality Assurance teams for designing validation cases in our simulation environment, Apollo Car Operations Team for algorithm validation on an autonomous vehicle.

\bibliographystyle{IEEEtran}
\bibliography{IEEEabrv,./refs}

\end{document}